\documentclass[times, review, 10pt]{elsarticle}

\usepackage{amssymb}
\usepackage{amsmath}
\usepackage[numbers]{natbib}
\usepackage{subcaption}
\usepackage{graphicx}
\usepackage{float}
\usepackage{comment}
\usepackage{url}

\journal{Nuclear Physics B}

\begin{document}

\begin{frontmatter}

\title{CapStARE: Capsule-based Sequential Architecture for Robust and Efficient Gaze Estimation}                      


\author[1]{Miren Samaniego Alonso\corref{cor1}}
\ead{miren.samaniego@ehu.eus}
\cortext[cor1]{Corresponding Author}

\author[1]{Elena Lazkano Ortega}

\author[1]{Igor Rodriguez Rodriguez}

\affiliation[1]{organization={Department of Computer Science and Artificial Intelligence, Faculty of Informatics, University of the Basque Country (UPV/EHU)},
                city={Donostia-San Sebastián},
                state={Gipuzkoa},
                country={Spain}}

\begin{abstract}
    Human gaze estimation is essential for applications such as human-computer interaction, social robotics, and assistive systems. However, achieving accurate, interpretable, and real-time performance in unconstrained environments remains challenging. Existing appearance-based methods often face trade-offs between spatial robustness, computational efficiency, and effective use of contextual information. To address this, we introduce CapStARE, a capsule-based architecture that combines a frozen ConvNeXt backbone for efficient feature extraction, capsule formation with attention-based routing for structured facial reasoning, and dual GRU decoders for lightweight sequential modeling over short-horizon observation windows. This design preserves interpretable part-whole facial relationships while improving prediction stability through local contextual consistency. Experimental results demonstrate strong performance on ETH-XGaze ($3.36^\circ$) and MPIIFaceGaze ($2.65^\circ$), while also generalizing competitively on Gaze360 ($9.06^\circ$), all with real-time inference ($<10$ ms). These findings suggest that the proposed method provides a practical and robust framework for appearance-based gaze estimation in real-world interactive environments. The related code and experimental results are publicly available at: \url{https://github.com/toukapy/capsStare}
\end{abstract}

\begin{highlights}
\item Capsule-based gaze estimation with lightweight sequential reasoning.
\item Structured facial representations improve robustness and interpretability.
\item Short-context modeling improves gaze stability and ambiguity resolution.
\item Competitive cross-dataset generalization across gaze benchmarks.
\item Real-time inference achieved with under 10 ms latency.
\end{highlights}


\end{frontmatter}

\section{Introduction}

Human gaze is a central component of human communication, continuously conveying cues about attention, intention, and social engagement during everyday interactions. From early child development, where gaze following supports joint attention, social learning, and language acquisition, to adulthood, it remains essential for coordinating behavior and interpreting communicative intent \citep{byersheinlein2021gaze,franchak2022beyond}. In artificial systems, accurate gaze estimation has become a fundamental capability for applications including assistive technologies, behavioral analysis, and human-robot interaction, where understanding a person's gaze direction can substantially improve responsiveness, safety, and interaction quality \citep{stower2021gaze,hriAdmoniHenny}. However, robust gaze estimation under unconstrained real-world conditions remains a significant challenge due to appearance variability, occlusions, illumination changes, and extreme head pose configurations. These challenges become particularly pronounced in unconstrained appearance-based gaze estimation, because models must infer subtle gaze-related facial cues directly from RGB imagery under highly variable environmental and behavioral conditions.

Despite substantial progress in appearance-based gaze estimation, current approaches still face a persistent trade-off between robustness, interpretability, and deployment efficiency. CNN-based architectures such as Full Face \citep{Zhang_2017} and L2CS-Net \citep{abdelrahman2022l2csnetfinegrainedgazeestimation} achieve strong spatial feature extraction, but their representations often remain largely implicit, limiting interpretability and reducing robustness under challenging visual conditions. To address some of these limitations, transformer-based and attention-driven approaches explored richer reasoning and long-range dependency modeling, but these gains frequently come at the cost of higher computational complexity and reduced real-time suitability \citep{9956687}. In parallel, sequential gaze estimation research suggests that incorporating neighboring observations can improve prediction stability beyond single-frame inference, particularly under ambiguous appearance conditions \citep{kellnhofer2019gaze360physicallyunconstrainedgaze,jindal2024spatiotemporalattentiongaussianprocesses}. However, many existing sequential approaches rely on computationally expensive spatiotemporal processing or monolithic recurrent pipelines that limit efficiency and flexibility in real-world deployment scenarios.

We argue that robust gaze estimation requires three complementary properties that are rarely addressed simultaneously: (1) structured spatial representations that preserve semantically meaningful facial organization, (2) efficient relational reasoning across localized facial regions, and (3) lightweight contextual modeling across neighboring observations. Existing approaches typically emphasize only one of these aspects, often sacrificing interpretability, efficiency, or deployment practicality.

Guided by this perspective, we present CapStARE (\textbf{Cap}sule-based \textbf{S}equen\textbf{t}ial \textbf{A}rchitecture for \textbf{R}obust and \textbf{E}fficient Gaze Estimation), a unified framework that combines capsule-based facial structuring, attention-driven relational routing, and lightweight sequential contextualization within a deployment-oriented architecture. Unlike conventional CNN embeddings, capsule representations preserve localized facial organization and semantically meaningful part-whole relationships. Self-attention routing then enables efficient relational aggregation across capsule tokens and neighboring observations without the computational burden of iterative capsule routing. Finally, dual GRU decoders incorporate short-horizon contextual consistency to improve prediction stability beyond single-frame inference while maintaining real-time efficiency.

Our contributions are summarized as follows:

\begin{enumerate}
    \item \textbf{Unified Structured Contextual Gaze Modeling: }We introduce a unified gaze estimation framework that combines structured capsule representations, attention-based relational routing, and lightweight sequential contextualization to jointly improve robustness, interpretability, and deployment efficiency.
    
    \item \textbf{Efficient Capsule Routing through Self-Attention: }We reformulate capsule interaction through self attention based routing, enabling scalable relational reasoning across localized facial regions and neighboring observations while avoiding the computational overhead of iterative routing mechanisms.
    
    \item \textbf{Lightweight Sequential Modeling: }We propose dual GRU-based sequential decoders for short-horizon ordered observation windows, improving prediction stability beyond single-frame inference while preserving computational efficiency.
    
    \item \textbf{Real-time Performance with Strong Generalization: }We demonstrate competitive performance across ETH-XGaze, MPIIFaceGaze, and Gaze360 while maintaining low latency and suitability for real-world interactive systems.

\end{enumerate}

\section{State of the art}

Human gaze estimation is a fundamental problem in computer vision with applications in human-computer interaction, assistive technologies, behavioral analysis, and robotics \citep{article}. Early approaches relied on infrared eye-tracking systems that estimate gaze through geoemtric relationships between pupil centers and corneal reflections \citep{duchowski2017eye}. Although highly accurate under controlled conditions, their dependence on specialized hardware and calibration procedures limits real-world deployment, motivating the development of appearance-based methods that infer gaze directly from RGB imagery.

The emergence of large-scale benchmarks accelerated progress in appearance-based gaze estimation. MPIIFaceGaze \citep{Zhang_2015} established an early person-independent benchmark, ETH-XGaze \citep{zhang_2020} expanded pose, gaze, and illumination diversity under a unified evaluation protocol, and Gaze360 \citep{kellnhofer2019gaze360physicallyunconstrainedgaze} extended evaluation toward unconstrained 3D gaze estimation in real-world environments.

Modern appearance-based gaze estimation initially focused on convolutional neural networks (CNNs). Methods such as iTracker \citep{krafka2016eyetracking}, FullFace \citep{Zhang_2017}, and L2CS-Net \citep{abdelrahman2022l2csnetfinegrainedgazeestimation} progressively evolved from eye-centered analysis toward full-face representations that jointly exploit ocular appearance, facial geometry, and head pose cues. Despite their strong performance, CNN-based approaches typically rely on implicit feature representations that provide limited interpretability and may be sensitive to challenging viewing conditions.

To improve structural reasoning, capsule networks were introduced to preserve hierarchical part-whole relationships through structured representations and routing mechanisms. In gaze estimation, GazeCaps \citep{10208362} and GazeCapsNet \citep{s25041224} adapted this paradigm to facial analysis, demonstrating that capsule-based representations can improve robustness while maintaining structured and semantically organized feature representations. While GazeCaps introduced self-attention-routed capsules for gaze estimation and GazeCapsNet focused on lightweight capsule-based inference, neither approach explicitly incorporates contextual modeling across neighboring observations. Nevertheless, capsule architectures remain relatively underexplored within modern gaze estimation pipelines.

Recent research has also explored the incorporation of geometric facial cues beyond appearance alone. For example, joint learning frameworks combining facial landmark detection and head-pose estimation have demonstrated that explicit geometric supervision can improve facial representation learning while preserving pose-related information \citep{ZOU2025111393}. Similarly, LGNet \citep{info17030224} introduced a landmark-guided gaze estimation framework that generates eye keypoints through a conditional variational autoencoder and integrates geometric priors with appearance features through cross-attention mechanisms, improving robustness under challenging conditions such as occlusions and low-quality visual inputs.

In parallel, transformer-based architectures have increasingly been adopted to improve contextual reasoning. GazeTR \citep{9956687} combined convolutional feature extraction with transformer-based contextual aggregation, while SwAT \citep{jindal2024spatiotemporalattentiongaussianprocesses} introduced spatiotemporal attention mechanisms for jointly modeling spatial and sequential dependencies. Lightweight alternatives such as PoolFormer \citep{app13105901}, Hybrid ViTNet \citep{karmi2025appearance}, Gaze-LLE \citep{ryan2025gaze}, and Eye-To-Action \citep{wang2026eye} further explored the trade-off between contextual modeling capability and computational efficiency. More recently, GMGaze \citep{ZHAO2026116120} explored semantic conditioning through CLIP representations, multiscale transformers, and domain adaptation mechanisms to improve contextual reasoning and cross-domain robustness.

Beyond architectural design, several recent studies have focused on improving representation quality and generalization. SAZE \citep{KIM2024110441} introduced stochastic subject-wise adversarial learning to reduce appearance bias and improve person-independent gaze estimation. ADGaze \citep{LI2025111536} proposed an anisotropic label distribution learning framework that combines coarse-to-fine estimation with structured angular supervision, demonstrating that explicitly modeling relationships between neighboring gaze directions can improve estimation accuracy. More recently, CGaG \citep{XIA2025111244} explored collaborative contrastive learning to learn domain-invariant gaze representations and improve cross-dataset transferability under heterogeneous acquisition conditions. GMMGaze \citep{ZHAO2026113452} further investigated distribution-aware gaze estimation through Gaussian Mixture Model-guided dynamic binning and coarse-to-fine prediction, showing that adaptive supervision based on the multimodal distribution of gaze angles can improve robustness under data imbalance and challenging visual conditions such as occlusion, blur, and illumination variations.

Sequential gaze estimation research has shown that incorporating neighboring observations improves robustness beyond single-frame inference. Approaches such as iTracker-BiLSTM \citep{8784770} and Gaze360 \citep{kellnhofer2019gaze360physicallyunconstrainedgaze} demonstrated the benefits of recurrent and spatiotemporal modeling, while more recent studies \citep{personnic2026learning,xia2025timegazer} suggest that local temporal consistency alone can substantially improve prediction stability. However, many sequential approaches still rely on computationally expensive spatiotemporal processing or complex recurrent architectures that limit deployment practicality.

Despite substantial progress across CNN-based, capsule-based, transformer-based, and sequential frameworks, several challenges remain unresolved. CNN-based systems often provide limited interpretability, capsule architectures remain underexplored, transformer-based methods may increase computational cost, and sequential models frequently reduce real-time suitability. Furthermore, existing capsule-based gaze estimation approaches have largely focused on improving spatial representation learning, while contextual information across neighboring observations remains relatively unexplored. Consequently, there remains a need for gaze estimation frameworks that jointly combine structured spatial representations, efficient contextual reasoning, and lightweight sequential modeling within a unified architecture. Unlike previous capsule-based gaze estimation methods, CapStARE combines capsule structuring with lightweight sequential modeling and attention-based contextual aggregation. This gap motivates the development of CapStARE.

\section{Interpretable Sequential Context Gaze Estimation}

The proposed CapStARE framework focuses on lightweight contextual aggregation across locally ordered observations to improve prediction stability beyond single-frame estimation. Rather than modeling unrestricted temporal dynamics, the architecture leverages short-horizon observation windows that preserve local appearance continuity while maintaining deployment efficiency.

These windows may correspond to adjacent video frames or protocol-defined neighboring observations constructed within appearance-based gaze datasets. Dataset-specific sequence construction protocols are detailed in Section~\ref{sec:experimentation}.

Unlike fully spatiotemporal gaze estimation pipelines, which rely on extended temporal processing, CapStARE exploits local contextual continuity across neighboring observations to stabilize predictions under ambiguous appearance conditions. This design reduces computational complexity and memory requirements while retaining the contextual information most relevant for appearance-based gaze estimation.

Building on CNN-based methods such as FullFace \citep{Zhang_2017} and L2CS-Net \citep{abdelrahman2022l2csnetfinegrainedgazeestimation}, CapStARE preserves strong facial representations through efficient convolutional encoding. Inspired by sequential frameworks such as Gaze360 \citep{kellnhofer2019gaze360physicallyunconstrainedgaze}, it incorporates lightweight sequential modeling while avoiding the computational cost and strong temporal assumptions of full video pipelines. In contrast to transformer-based approaches \citep{ryan2025gaze,app13105901,9956687}, the proposed architecture combines self-attention with capsule structuring to preserve localized facial correspondence and support interpretable part-whole relational reasoning.

\begin{figure}[t]
  \centering
  \includegraphics[width=\linewidth]{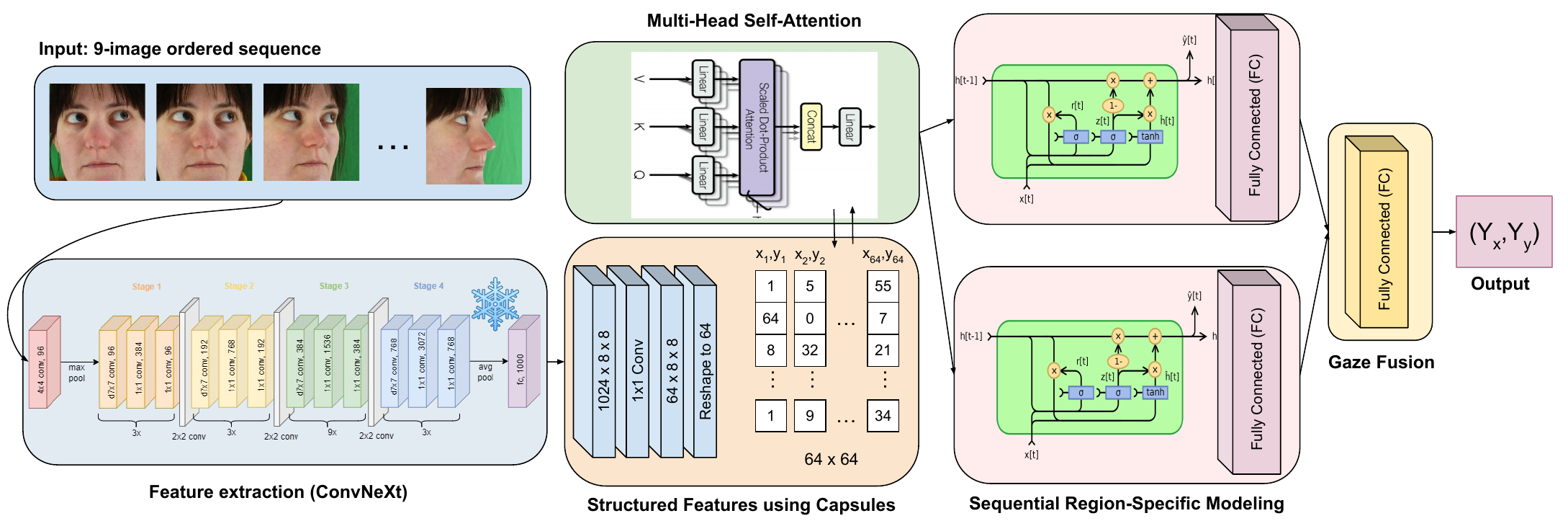}
  \caption{\textbf{Overview of CapStARE.} Consecutive observations within a short-horizon ordered window are independently processed by a frozen ConvNeXt encoder, transformed into capsule tokens, routed through joint spatial-sequential self-attention, and decoded by two independent GRU branches before final gaze fusion.}
\label{fig:firstApproachScheme}
\end{figure}

Single-frame CNN-based gaze estimation provides strong spatial feature extraction but cannot explicitly model dependencies across neighboring observations, while transformer-based and fully spatiotemporal approaches often increase computational cost and hinder real-time deployment. CapStARE addresses this trade-off through a modular architecture that combines structured facial decomposition with efficient contextual aggregation.

The framework consists of four stages: convolutional feature extraction, capsule-based facial decomposition, joint spatial-contextual routing, and dual recurrent decoding with late fusion. Rather than imposing hand-crafted assumptions on decoder behavior, independent optimization allows complementary recurrent dynamics to emerge naturally during training.

Figure~\ref{fig:firstApproachScheme} provides an overview of the proposed architecture, while the following subsections describe each component in detail.
\subsection{Feature Extraction}

For each observation within an ordered window, CapStARE first applies a frozen ConvNeXt encoder to extract robust facial representations. Freezing the encoder serves two complementary purposes: it reduces overfitting on relatively limited gaze datasets and decouples generic visual representation learning from downstream structured sequential reasoning. ConvNeXt \citep{liu2022convnet2020s} is selected due to its balance of hierarchical convolution, large receptive fields, and modern optimization design, enabling effective transfer from pretrained visual knowledge.

Formally, given an ordered observation window of length $T$, each frame $x_t$ of size $W \times H$ is processed independently as:
\begin{equation}
  F_t = E(x_t), \quad t \in \{1, ..., T\}
\end{equation}
where $E(\cdot)$ denotes the frozen ConvNeXt encoder and $F_t \in \mathbb{R}^{C \times H \times W}$ is the resulting feature map. Stacking all observations yields:
\begin{equation}
  \mathbf{F} \in \mathbb{R}^{B \times T \times C \times H \times W}
\end{equation}
where $C$ is the number of channels, $B$ is the batch size, and $T$ is the number of observations. This design preserves consistent spatial representations before relational aggregation across neighboring observations.

\subsection{Structured Feature Representation via Capsules}

Although convolutional feature grids are highly expressive, they are difficult to interpret directly in terms of semantically meaningful facial regions. To address this limitation, each spatial feature map is transformed into capsule tokens representing localized facial entities, such as eye regions, nose bridge, or facial contours, preserving coherent spatial organization.

For each encoded observation:
\begin{equation}
  C_{caps} = \phi(F_t) \in \mathbb{R}^{N \times D}
\end{equation}
where $N = H \times W$ denotes flattened spatial locations, $D$ is the capsule dimensionality, and $\phi(\cdot)$ is a learnable linear projection. Across the full ordered window:
\begin{equation}
  \mathbf{C}_{caps} \in \mathbb{R}^{B \times T \times N \times D}
\end{equation}

This representation preserves token-level correspondence with localized facial regions, improving interpretability and creating an organized representation space for relational routing. Unlike conventional dense convolutional embeddings, capsule tokens preserve explicit correspondence between latent representations and localized facial regions. This property is particularly important in gaze estimation, where subtle geometric variations in periocular appearance, eyelid configuration, eyebrow orientation, and broader facial structure jointly contribute to gaze perception. By preserving localized structural organization, capsule representations facilitate more interpretable relational reasoning while improving robustness to viewpoint variation and partial occlusion.

Moreover, the capsule formulation naturally supports relational aggregation across neighboring observations. Since semantically related facial regions remain consistently represented across locally ordered sequences, contextual interactions can be modeled directly at the token level rather than through globally entangled feature embeddings. This structured representation therefore provides a stable foundation for lightweight contextual reasoning across short observation windows.

\subsection{Joint Spatial-Contextual Self-Attention}

Traditional routing-by-agreement in capsule networks is computationally iterative and difficult to scale efficiently. CapStARE instead reformulates routing as global relational weighting through multi-head self-attention, enabling capsules to exchange information across both spatial structure and adjacent observations.

First, the capsule tensor is reshaped into a unified token sequence:
\begin{equation}
  \tilde{C}_{caps} \in \mathbb{R}^{B \times (T \cdot N) \times D}
\end{equation}
Projection into query, key, and value spaces then yields:
\begin{equation}
    Q = \tilde{C}_{caps}W_Q,\quad K = \tilde{C}_{caps}W_K,\quad V = \tilde{C}_{caps}W_V
\end{equation}
and attention-based routing is computed as:
\begin{gather}
  A = \text{softmax}\left(\frac{QK^T}{\sqrt{D}}\right), \quad C_{out} = AV
\end{gather}
where $W_Q$, $W_K$, and $W_V$ are learnable projections. This mechanism allows capsules associated with specific facial regions at observation step $t$ to directly interact with semantically related capsules from adjacent observation steps, thereby stabilizing predictions under subtle pose variation or appearance changes.

Importantly, this formulation replaces iterative routing-by-agreement with fully differentiable relational aggregation through attention weighting. This substantially improves scalability and computational efficiency while preserving the core objective of capsule interaction: dynamically strengthening relationships between semantically coherent entities. Unlike transformer-only approaches that operate over globally mixed embeddings, the proposed routing mechanism preserves token-level correspondence with localized facial structure throughout contextual aggregation.

Additionally, routing jointly across spatial and contextual dimensions enables the architecture to model both intra-observation and inter-observation dependencies within a unified attention space. This allows facial regions exhibiting stable gaze-related structure across neighboring observations to reinforce one another during contextual reasoning, thereby improving robustness under subtle appearance variation, transient occlusion, or ambiguous head pose configurations.

\subsection{Dual Sequential Decoding}

Following relational routing, the structured sequence is decoded by two parallel GRU branches:
\begin{equation}
  h_k = \text{GRU}_k(C_{out}), \quad y_k = W_k h_k + b_k, \quad k \in \{1,2\}
\end{equation}
where each decoder maintains independent parameters. 

The purpose of this design is not to enforce predefined movement categories, but rather to increase representational diversity by allowing independent recurrent pathways to emerge during optimization. This avoids compressing diverse contextual cues into a single hidden representation and supports empirical specialization through independent gradient flow.

This design is motivated by the observation that contextual gaze dynamics may emerge at multiple representational scales simultaneously. Different recurrent pathways may therefore specialize in distinct contextual dependencies, appearance transitions, or local structural dynamics without requiring explicit hand-crafted decomposition strategies. By preserving decoder independence prior to fusion, CapStARE increases representational flexibility while avoiding premature compression of contextual information into a single recurrent state.

Moreover, lightweight GRU-based decoding provides an effective balance between contextual modeling capacity and computational efficiency. Compared to heavier transformer-based sequential architectures, recurrent decoding maintains substantially lower memory requirements and inference latency, making the proposed framework more suitable for deployment-oriented gaze estimation scenarios.

\subsection{Late Gaze Fusion}

The outputs of the two decoders branches are integrated only after independent sequential reasoning:
\begin{equation}
  y_{concat} = \text{concat}(y_1, y_2), \quad \hat{y} = W_f y_{concat} + b_f
\end{equation}
Late fusion preserves branch independence while enabling adaptive integration of complementary contextual evidence.

Overall, CapStARE unifies structured facial representation, lightweight contextual aggregation, and efficient sequential reasoning within a deployment-oriented gaze estimation framework. 

\section{Experimentation}
\label{sec:experimentation}

To rigorously evaluate CapStARE and address both predictive effectiveness and methodological validity, our experimental framework is organized around four complementary objectives: (1) establishing reproducible dataset usage and evaluation protocols, (2) quantifying predictive performance through standardized gaze estimation metrics, (3) analyzing computational efficiency and deployment feasibility, and (4) systematically investigating architectural design choices through controlled ablation studies. Collectively, these experiments are designed not only to evaluate predictive accuracy, but also to validate whether CapStARE's short-horizon sequential modeling offers meaningful advantages over static appearance-based alternatives without overstating temporal assumptions.

\subsection{Dataset Usage Protocol and Evaluation Settings}

CapStARE is designed to exploit short-horizon contextual consistency in appearance-based gaze estimation while remaining compatible with datasets that were not originally designed as continuous video benchmarks. Rather than modeling long-term temporal dynamics, the framework leverages locally ordered observations to improve prediction stability.

\textbf{Primary training dataset.} ETH-XGaze is used for training, validation, hyperparameter tuning, backbone comparisons, and architectural ablations due to its scale, subject diversity, and variability in gaze direction, head pose, illumination, and appearance.

\textbf{Cross-dataset generalization.} To evaluate robustness beyond the source domain, the final ETH-XGaze trained model is tested on MPIIFaceGaze and Gaze360 without retraining. Results are reported using source-to-target notation:
\begin{equation}
  D_E \rightarrow D_M,\quad D_E \rightarrow D_G
\end{equation}
where $D_E$, $D_M$ and $D_G$ denote ETH-XGaze, MPIIFaceGaze and Gaze360 datasets, respectively. 

\begin{figure}[t]
  \centering
  \includegraphics[width=\linewidth]{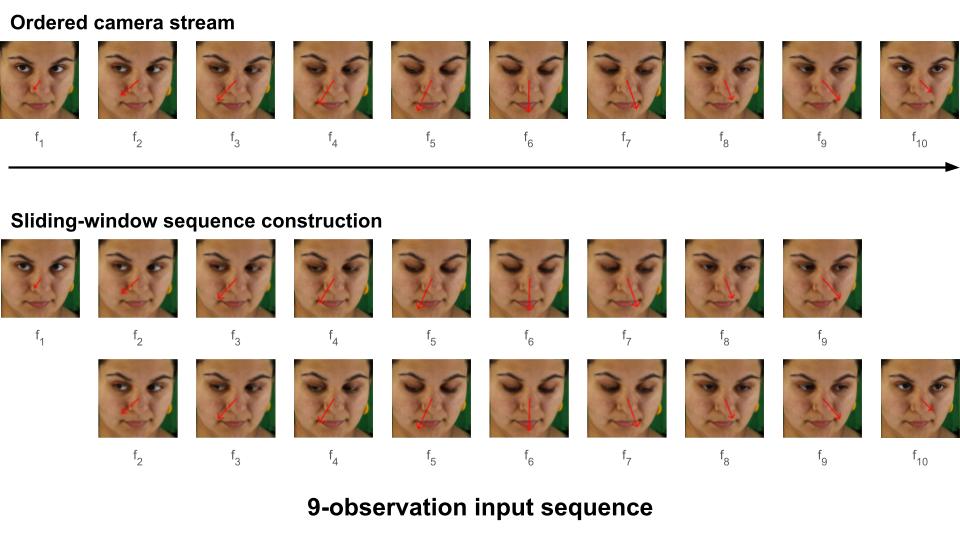}
  \caption{Sliding-window sequence construction used in CapStARE. Ordered neighboring observations from the same camera stream are grouped into fixed-length 9-observation sequences, preserving short-horizon gaze and appearance continuity.}
\label{fig:sequenceCons}
\end{figure}

\textbf{Sequence construction rationale. }Sequences are generated independently within each camera stream using fixed-length sliding windows of nine observations and stride 1 (Figure \ref{fig:sequenceCons}). Neighboring samples therefore preserve local continuity in appearance and gaze direction while avoiding artificial aggregation across different viewpoints. Although ETH-XGaze is not a video benchmark, its ordered acquisition process provides sufficient local consistency for evaluating short-horizon contextual reasoning.

All experiments follow the official subject-disjoint protocol (65/16/15 subjects for training, validation, and testing). Dataset partitioning is performed before sequence construction, and sequences are generated independently within each split and camera stream, preventing subject-level and temporal leakage.

Gaze prediction is performed for the central observation of each sequence. Border replication is used when insufficient context is available near sequence boundaries, ensuring fixed-length inputs throughout training and evaluation. Under this protocol, the training and validation splits contain approximately 605K and 150K sequences, respectively.

Since border-replication padding is applied during sequence construction, each observation contributes to exactly one centered temporal sequence, preserving the original dataset cardinality while maintaining temporal continuity.

\subsection{Performance Metrics}

To evaluate predictive performance, we adopt angular error as the primary metric, as it remains the most widely used and perceptually meaningful measure for appearance-based gaze estimation. Although CapStARE predicts 2D gaze coordinates, predictions are converted into normalized 3D gaze direction vectors under a pinhole camera assumption to enable direct directional comparison with ground truth.

Angular error is computed as:
\begin{equation}
  \theta = \arccos\left(\mathbf{p}_{3D} \cdot \mathbf{g}_{3D}\right)\times\frac{180}{\pi}
\end{equation}
where $\mathbf{p}_{3D}$ and $\mathbf{g}_{3D}$ denote normalized predicted and ground truth 3D gaze vectors respectively. Lower values indicate more accurate directional estimation.

During optimization, Mean Squared Error (MSE) loss is applied to predicted gaze coordinates; angular error is reserved for optimization stability from interpretable final performance assessment.

\subsection{Data Preprocessing and Normalization}

To ensure methodological consistency across datasets, all observations undergo standardized preprocessing prior to training and evaluation. Full-face images are used instead of isolated eye crops in order to preserve global facial context, including head pose, facial orientation, and broader appearance cues relevant for gaze estimation. Images are resized to the ConvNeXt input resolution and normalized using ImageNet mean and standard deviation statistics to maintain compatibility with pretrained backbone initialization and stable feature extraction.

Preprocessing is applied independently to each observation before sequence construction. After normalization, fixed-length ordered windows of nine observations are generated according to the dataset-specific protocol described in Figure~\ref{fig:sequenceCons}. This separation between per-frame preprocessing and sequence formation preserves consistent input normalization while enabling short-horizon contextual modeling across neighboring observations.

\subsection{Computational Cost}

To assess deployment feasibility, we report FLOPs, inference time, and trainable parameter count. FLOPs provide a hardware-independent estimate of arithmetic complexity, inference time reflects practical latency under deployment conditions, and parameter count captures model size and memory requirements.

These metrics are particularly relevant because CapStARE introduces structured routing and dual recurrent decoding beyond standard single-frame CNN baselines. Reporting computational cost therefore allows direct examination of whether performance gains are achieved through efficient structured reasoning rather than impractical architectural expansion.

\subsection{Model Analysis and Ablation Study}

To comprehensively validate CapStARE's design, we perform progressive architectural analysis across four experimental axes.

First, internal configuration studies evaluate capsule count, capsule dimensionality, attention head count, and GRU hidden size to identify efficient structural trade-offs.

Second, backbone substitution experiments replace ConvNeXt with EfficientNet, ViT, and Swin Transformer while preserving downstream modules. This isolates the influence of feature extraction quality from sequential and capsule reasoning. Where transformer backbones underperform, results are interpreted in the context of dataset scale, inductive bias, and pretrained transfer suitability rather than raw architectural popularity.

Third, structural ablations selectively remove or modify major components, including capsules, self-attention routing, single versus dual GRU decoding, and weight sharing. These experiments explicitly determine whether observed gains arise from structured capsules, ordered routing, or decoder diversity.

Fourth, sequence sensitivity experiments compare single-observation input against structured multi-observation windows in order to quantify the contribution of short-horizon contextual reasoning beyond static appearance estimation.

Results from these studies are presented in Section~\ref{sec:results}.

\subsection{Experimental Setup}

All experiments were conducted on a Linux workstation equipped with an NVIDIA RTX 4070Ti GPU, Intel Core i7-14700KF CPU, and 32 GB RAM using Python 3.12, Pytorch 2.6.0, Torchvision 0.21.0, and CUDA 12.4.

Models were trained for 30 epochs using Adam with learning rate $1\times10^{-5}$, weight decay $1\times10^{-5}$, cosine annealing, batch size 32 for training, and batch size 16 for validation. Dropout and L2 regularization were applied after capsule formation. ConvNeXt-Base was initialized from ImageNet pretrained weights and kept fully frozen after comparative partial-unfreezing trials demonstrated superior stability and efficiency.

Training efficiency was further improved through mixed precision (FP16/FP32), TorchDynamo compilation, cuDNN autotuning, and Accelerate-based hardware management.

This setup reflects a deliberate design philosophy: CapStARE is intended to improve robustness and interpretability primarily through structured ordered reasoning rather than through excessive backbone finetuning or computational scale alone.

\section{Results}
\label{sec:results}

To evaluate CapStARE under protocol-consistent and deployment-oriented conditions, experiments are organized into three complementary analyses: (1) internal configuration studies, which identify efficient architectural operating points under a fixed ETH-XGaze training and validation protocol; (2) controlled ablation studies, which isolate the contribution of structured capsules, relational routing, and dual-path sequential decoding; and (3) comparative benchmarking, where within-dataset and cross-dataset evaluations are explicitly separated to avoid protocol conflation. Beyond absolute predictive accuracy, experiments additionally examine how structured contextual aggregation influences robustness, representational efficiency, and cross-domain transfer behavior under heterogeneous appearance conditions.

\subsection{Internal Configuration Study}

We first evaluate the influence of capsule count and attention head count under the standardized ETH-XGaze training and validation protocol. Four configurations were trained under identical conditions.

Table \ref{tab:changeHeads} shows that the 4-capsule/4-head configuration achieved the lowest angular error ($3.36^\circ$) while requiring the smallest parameter count (13M). Increasing either the number of capsules or attention heads does not improve performance despite the additional model capacity.

\begin{table}[h]
\centering

\begin{tabular}{l l l l l l}
    \hline
    \textbf{Caps} & 
    \textbf{Heads} & 
    \textbf{Err (°)} & 
    \textbf{Size} & 
    \textbf{Infer (ms)} & 
    \textbf{FLOPs (M)} \\
    \hline
    \textbf{4} & \textbf{4} & \textbf{3.36} & \textbf{13.00M} & \textbf{8.226} & \textbf{138.56} \\
    4 & 8 & 3.94 & 13.01M & 8.347 & 138.56 \\
    8 & 4 & 3.95 & 25.85M & 8.382 & 138.68 \\
    8 & 8 & 3.93 & 25.97M & 8.509 & 138.68 \\
    \hline
\end{tabular}

\caption{Performance comparison of different capsule and attention head configurations in a ConvNeXt-based gaze estimation model.}
\label{tab:changeHeads}
\end{table}

\begin{figure}[h]
    \centering
    \includegraphics[width=0.8\linewidth]{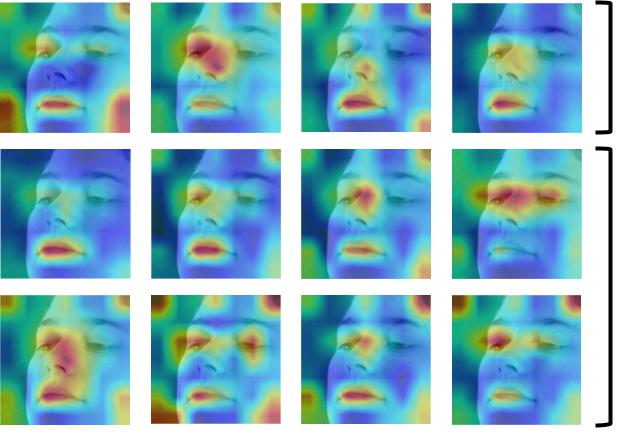}
    \caption{Comparison of feature representations with 4 vs. 8 capsules. The 4-capsule configuration (first row) focuses on distinct facial regions, while 8 capsules (second and third rows) provide finer but often redundant decompositions.}
    \label{fig:capsuleComparison}
\end{figure}

\begin{figure}[h!]
    \centering
    \includegraphics[width=0.95\linewidth]{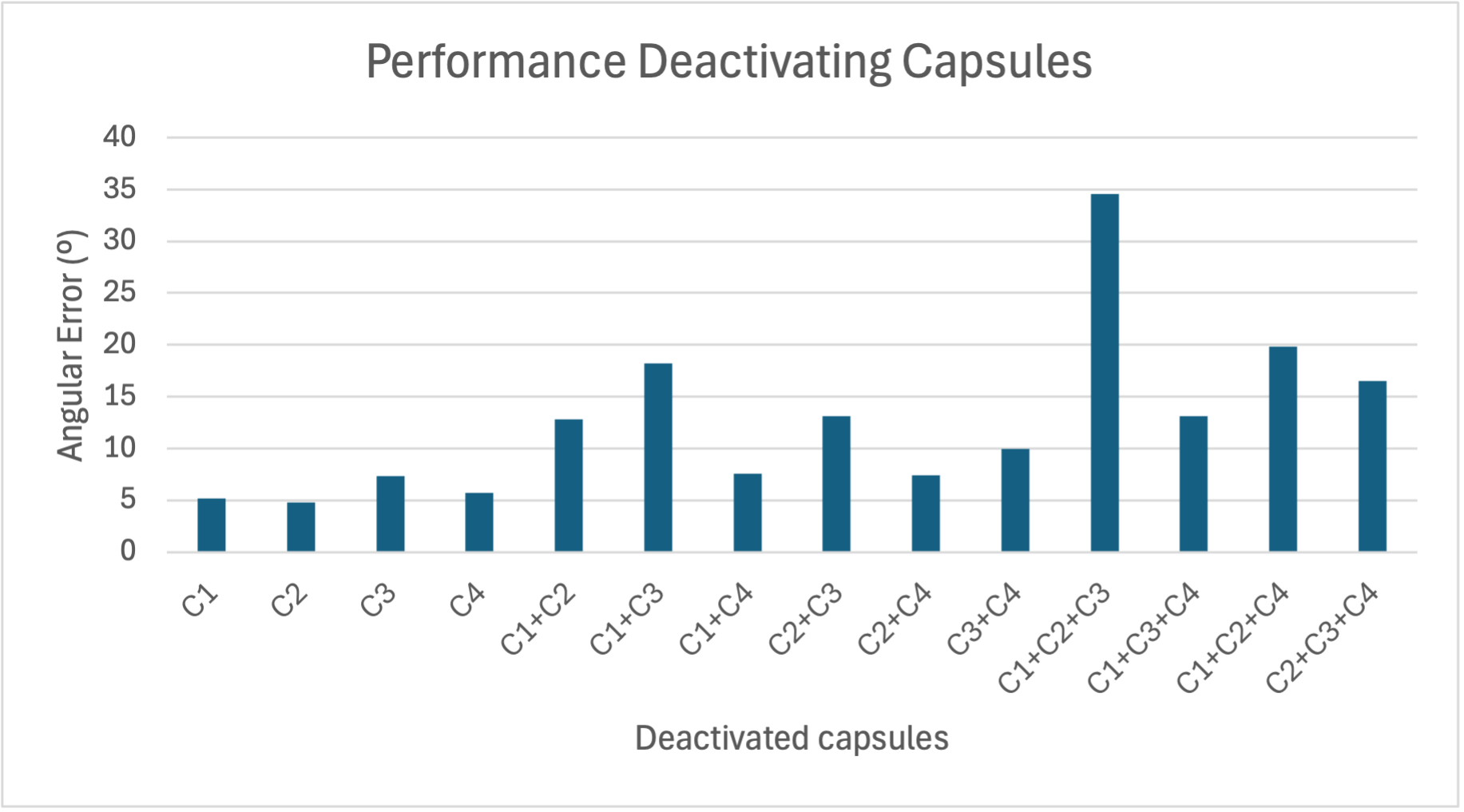}
\caption{Effect of capsule deactivation. Horizontal axis consists of groups of deactivated capsules.}
\label{fig:removeSomeCaps}
\end{figure}

Figure \ref{fig:capsuleComparison} provides a qualitative comparison between the 4-capsule and 8-capsule configurations. While both settings produce localized facial representations, the 8-capsule configuration exhibits a final decomposition with more overlapping activations.

Accordingly, the 4-capsule/4-head configuration was selected as the default architecture for all subsequent experiments.

Figure \ref{fig:removeSomeCaps} presents a capsule knockout analysis. Removing any individual capsule increases the angular error relative to the full model, indicating that all learned capsule representations contribute to the final prediction. The largest degradation among single-capsule removals is observed when Capsule 3 is deactivated. Performance deteriorates further when multiple capsules are removed simultaneously, with the highest error occurring when only one capsule remains active. These results indicate that accurate gaze estimation benefits from the joint contribution of multiple capsule representations. 

Building on the selected structural configuration, we next evaluate the effect of capsule and GRU hidden dimensionality.

\begin{table}[h]
\centering

\begin{tabular}{l l l l l l}
\hline
\textbf{Caps Dim} & 
\textbf{Hid Dim} & 
\textbf{Err (°)} & 
\textbf{Size (M)} & 
\textbf{Infer (ms)} & 
\textbf{FLOPs (M)} \\
\hline
64 & 64 & 4.17 & 12.94M & 8.064 & 138.56 \\
\textbf{64} & \textbf{128} & \textbf{3.36} & \textbf{13.00M} & \textbf{8.226} & \textbf{138.56} \\
64 & 256 & 3.61 & 13.4M & 5.810 & 138.56 \\
128 & 128 & 3.41 & 26M & 8.392 & 138.68 \\
\textbf{128} & \textbf{256} & \textbf{3.32} & \textbf{26.4M} & \textbf{5.865} & \textbf{138.68} \\
256 & 256 & 3.36 & 52.75M & 5.879 & 138.92 \\
\hline
\end{tabular}

\caption{Effect of capsule and hidden layer dimensionality on performance. Configurations range from 64D to 256D for capsules and hidden layers.}
\label{tab:dimensions}
\end{table}

Table \ref{tab:dimensions} shows relatively stable performance across dimensional scales. Although the 128D/256D configuration achieves the lowest angular error ($3.32^\circ$), the 64D/128D setting provides nearly identical performance ($3.36^\circ$) while requiring approximately half the parameter count (13M vs. 26.4M).

Given the negligible accuracy difference relative to the substantial increase in model size, the 64D/128D configuration was selected as the default setting for subsequent experiments.

We next evaluate alternative backbone architectures while preserving all downstream CapStARE components.

\begin{table}[h]
\centering

\begin{tabular}{l l l l}
\hline
\textbf{Backbone} & 
\textbf{Err (°)} & 
\textbf{Infer (ms)} & 
\textbf{FLOPs (M)} \\
\hline
EfficientNet & 7.16 & 2.053 & 14.29 \\
\textbf{ConvNeXT} & \textbf{3.36} & \textbf{8.226} & \textbf{138.56} \\
ViT & 6.26 & 10.617 & 101.94 \\
Swin Transformer & 10.30 & 9.813 & 139.20 \\
\hline
\end{tabular}

\caption{Backbone comparison for gaze estimation. ConvNeXt achieves the lowest error ($3.36^\circ$) while meeting real-time inference requirements.}
\label{tab:encoderChange}
\end{table}

Table \ref{tab:encoderChange} shows that ConvNeXt achieves the lowest angular error ($3.36^\circ$) among the evaluated backbones while remaining compatible with real-time inference. EfficientNet provides lower latency but substantially higher error, whereas ViT and Swin Transformer underperform under the evaluated protocol.

Consequently, ConvNeXt was selected as the default backbone for all subsequent experiments.

\subsection{Ablation Study}

To isolate the contribution of structured capsules and self-attention routing, we selectively disable each component while preserving all other settings.

\begin{table}[h]
\centering

\begin{tabular}{l l l l l l}
\hline
\textbf{Caps} & 
\textbf{Attn} & 
\textbf{Err (°)} & 
\textbf{Size (M)} & 
\textbf{Infer (ms)} & 
\textbf{FLOPs (M)} \\
\hline
\textbf{Yes} & \textbf{Yes} & \textbf{3.36} & \textbf{13.00} & \textbf{8.266} & \textbf{138.68} \\
No & Yes & 4.36 & 3.38 & 6.078 & 138.50 \\
Yes & No & 3.84 & 12.9 & 5.776 & 138.50 \\
No & No & 8.19 & 38.7 & 6.147 & 138.44 \\
\hline
\end{tabular}

\caption{Ablation study of capsule and attention components. Results show their individual and combined contributions to gaze estimation performance.}
\label{tab:noCapsulenoAtt}
\end{table}

The full CapStARE configuration achieves the strongest performance ($3.36^\circ$). Removing capsules while retaining attention increases the error to $4.36^\circ$, whereas removing attention while preserving capsules results in $3.84^\circ$. The largest degradation is observed when both components are removed simultaneously, reaching $8.19^\circ$.

These results indicate that both capsule decomposition and attention routing contribute to the final performance, with the complete architecture providing the strongest overall results.

We next evaluate decoder architecture by comparing dual-path decoding against a single shared decoder.

\begin{table}[h]
\centering
\begin{tabular}{l l l l}
\hline
\textbf{Decoder} & 
\textbf{Err (°)} & 
\textbf{Infer (ms)} & 
\textbf{FLOPs (M)} \\
\hline
\textbf{Double} & \textbf{3.32} & \textbf{5.865} & \textbf{138.68} \\
Single & 7.18 & 5.739 & 138.92 \\
\hline
\end{tabular}

\caption{Ablation of decoder architecture. Comparison between dual independent GRUs and a single shared decoder.}
\label{tab:changeDecoder}
\end{table}

Table \ref{tab:changeDecoder} shows that the dual-decoder configuration substantially outperforms the single-decoder alternative, reducing the angular error from $7.18^\circ$ to $3.32^\circ$ with only minor differences in computational cost.

Similarly, Table \ref{tab:changeWeights} evaluates the effect of weight sharing between decoder branches. Sharing parameters degrades performance from $3.32^\circ$ to $6.14^\circ$, despite nearly identical inference cost and FLOPs.

\begin{table}[h]
\centering

\begin{tabular}{l l l l}
\hline
\textbf{Weight Sharing} & 
\textbf{Err (°)} & 
\textbf{Infer (ms)} & 
\textbf{FLOPs (M)} \\
\hline
\textbf{No} & \textbf{3.32} & \textbf{5.865} & \textbf{138.68} \\
Yes & 6.41 & 5.972 & 138.92 \\
\hline
\end{tabular}

\caption{Effect of weight sharing in dual decoders. Independent parameters enable specialization across facial regions.}
\label{tab:changeWeights}
\end{table}

Taken together, these results suggest that both dual-path decoding and independent branch parameterization contribute to the final performance.

Sequence sensitivity experiments further compare single-observation and multi-observation input to quantify the contribution of short-horizon contextual modeling.

\begin{table}[h]
\centering

\begin{tabular}{l l l}
\hline
\textbf{Sequence Dim.} & 
\textbf{Err. ($^\circ$)} & 
\textbf{Infer. (ms)} \\
\hline
1 & 5.91 & 1.338 \\
3 & 4.71 & 3.442 \\
5 & 4.37 & 5.098 \\
7 & 3.82 & 8.013 \\
\textbf{9} & \textbf{3.36} & \textbf{8.266} \\
\hline
\end{tabular}

\caption{Effect of sequence length on gaze estimation performance. Different ordered observation window sizes were evaluated to analyze the impact of short-horizon contextual modeling on angular error and inference latency.}
 \label{tab:seqLen}
\end{table}

Table \ref{tab:seqLen} shows that performance improves consistently as sequence length increases, reducing the angular error from $5.91^\circ$ for single-observation input to $3.36^\circ$ for length-9 sequences. Accordingly, all subsequent experiments use a sequence length of 9 observations.

\begin{figure}[t]
\centering
\begin{subfigure}{0.19\textwidth} 
    \centering
    \includegraphics[width=\textwidth]{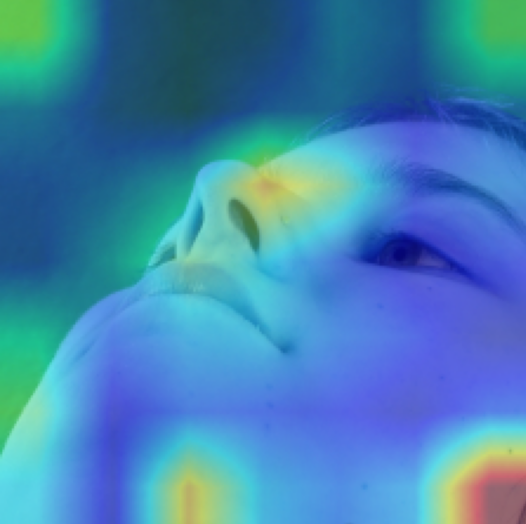}
    \label{fig:cap041000}
\end{subfigure}
\begin{subfigure}{0.19\textwidth}  
    \centering
    \includegraphics[width=\textwidth]{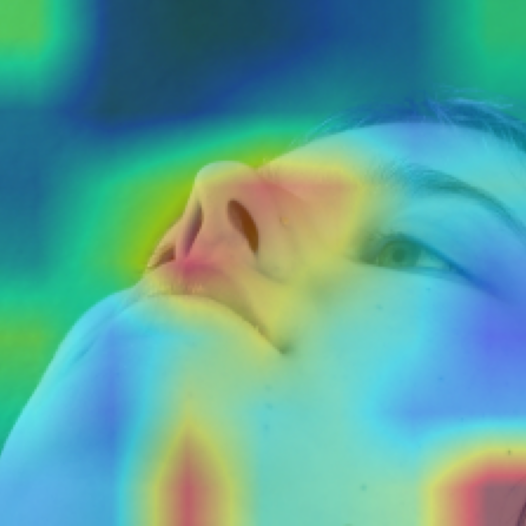}
    \label{fig:cap141000}
\end{subfigure}
\begin{subfigure}{0.19\textwidth}  
    \centering
    \includegraphics[width=\textwidth]{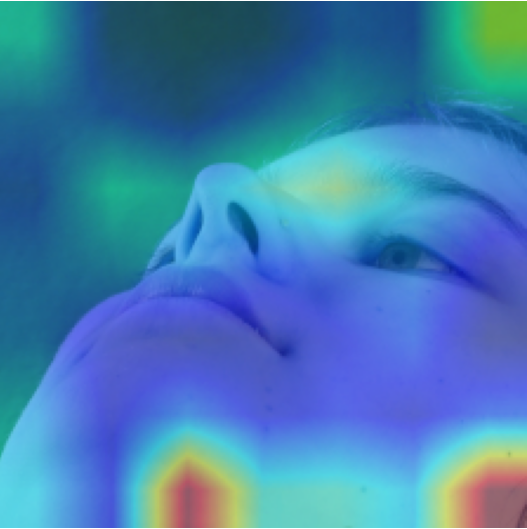}
    \label{fig:cap241000}
\end{subfigure}
\begin{subfigure}{0.19\textwidth}  
    \centering
    \includegraphics[width=\textwidth]{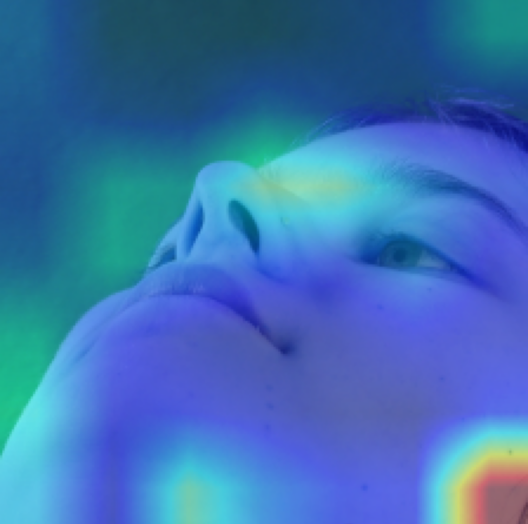}
    \label{fig:cap341000}
\end{subfigure}
\begin{subfigure}{0.19\textwidth}  
    \centering
    \includegraphics[width=\textwidth]{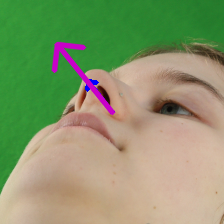}
    \label{fig:arrow1000}
\end{subfigure}
\begin{subfigure}{0.19\textwidth}  
    \centering
    \includegraphics[width=\textwidth]{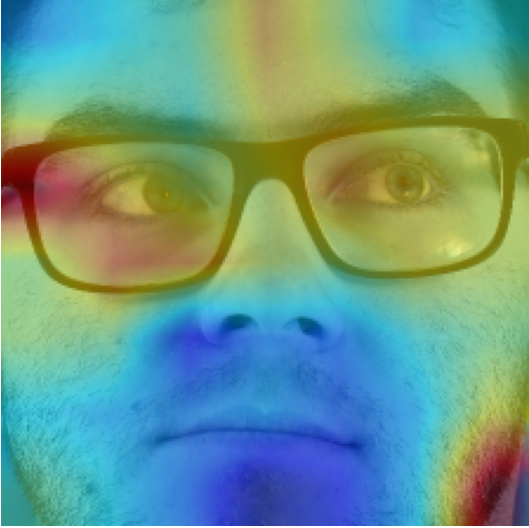}
    \label{fig:cap0410500}
\end{subfigure}
\begin{subfigure}{0.19\textwidth}  
    \centering
    \includegraphics[width=\textwidth]{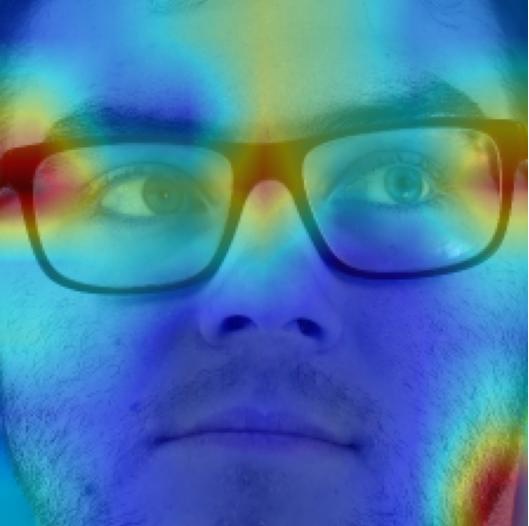}
    \label{fig:cap1410500}
\end{subfigure}
\begin{subfigure}{0.19\textwidth}  
    \centering
    \includegraphics[width=\textwidth]{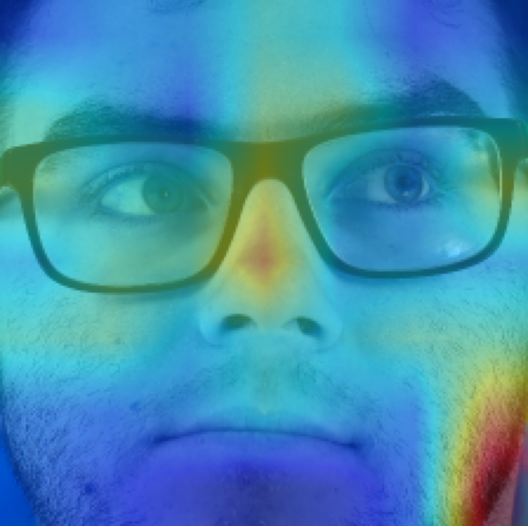}
    \label{fig:cap2410500}
\end{subfigure}
\begin{subfigure}{0.19\textwidth}  
    \centering
    \includegraphics[width=\textwidth]{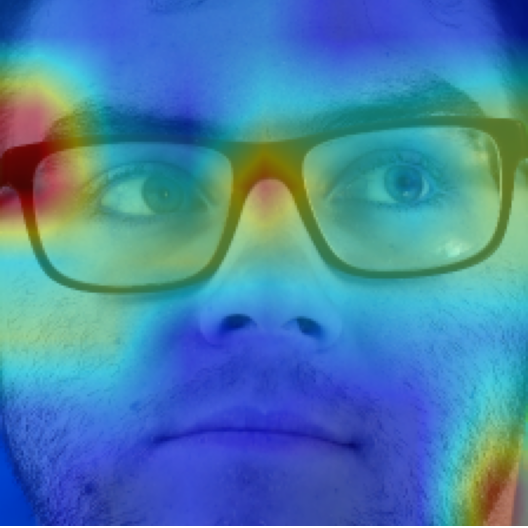}
    \label{fig:cap3410500}
\end{subfigure}
\begin{subfigure}{0.19\textwidth}  
    \centering
    \includegraphics[width=\textwidth]{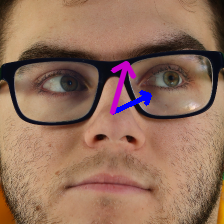}
    \label{fig:arrow10500}
\end{subfigure}
\begin{subfigure}{0.19\textwidth}  
    \centering
    \includegraphics[width=\textwidth]{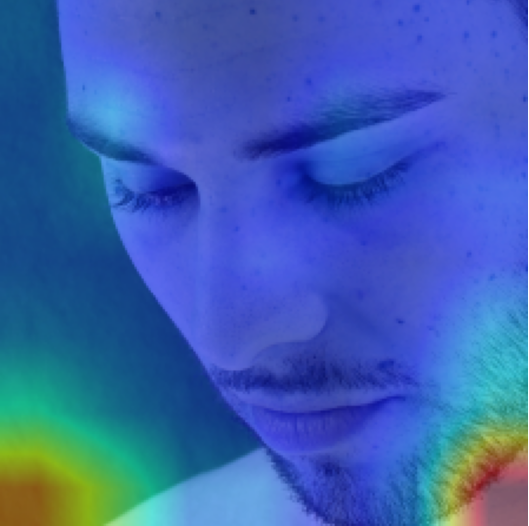}
    \label{fig:cap0423000}
\end{subfigure}
\begin{subfigure}{0.19\textwidth}  
    \centering
    \includegraphics[width=\textwidth]{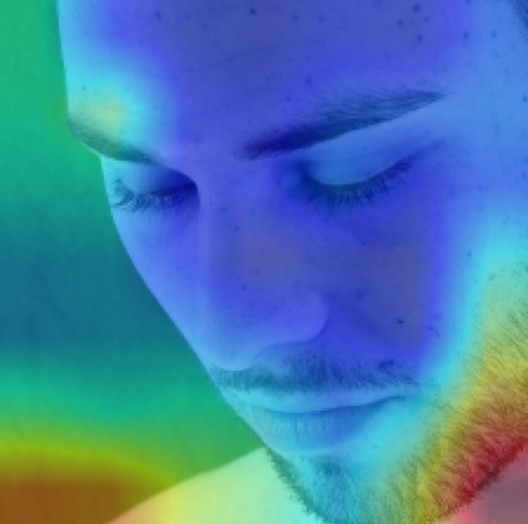}
    \label{fig:cap1423000}
\end{subfigure}
\begin{subfigure}{0.19\textwidth}  
    \centering
    \includegraphics[width=\textwidth]{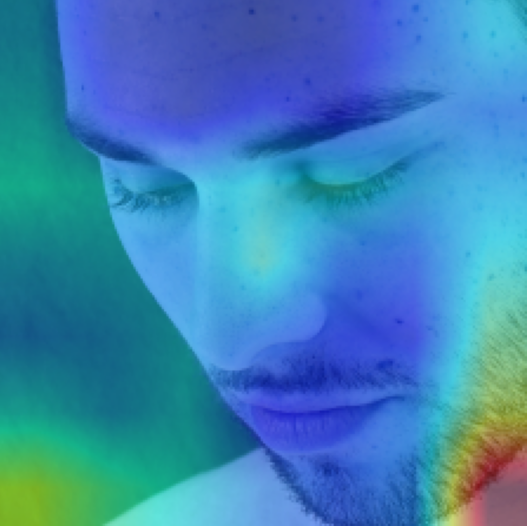}
    \label{fig:cap2423000}
\end{subfigure}
\begin{subfigure}{0.19\textwidth}  
    \centering
    \includegraphics[width=\textwidth]{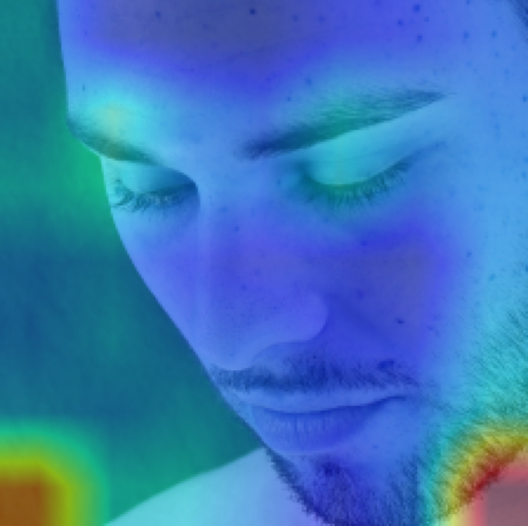}
    \label{fig:cap3423000}
\end{subfigure}
\begin{subfigure}{0.19\textwidth}  
    \centering
    \includegraphics[width=\textwidth]{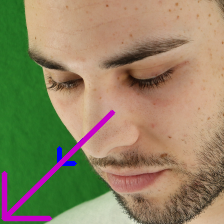}
    \label{fig:arrow23000}
\end{subfigure}
\caption{Learned capsule attention and gaze prediction. Each row shows: (1-4) capsule heatmaps highlighting specialized regions, and (5) predicted (purple) vs. ground truth (blue) gaze.}
\label{fig:capsuleSize}
\end{figure}

Figure \ref{fig:capsuleSize} illustrates representative capsule activations and corresponding gaze predictions. The learned capsules consistently focus on localized facial regions without explicit region supervision, while maintaining accurate gaze estimation under diverse appearance conditions.

\subsection{Comparative Benchmarking}

To evaluate CapStARE under both within-domain and cross-domain conditions, benchmarking experiments were separated. Within-dataset evaluation follows the standard ETH-XGaze subject-independent protocol, while cross-dataset evaluation assesses zero-shot transfer by training exclusively on ETH-XGaze and testing on external datasets without target-domain fine-tuning.

For cross-dataset evaluation, we report the results available in the original publications under ETH-XGaze-to-target transfer settings. Although minor implementation and preprocessing differences may exist, all methods were evaluated without target-domain fine-tuning.

\begin{table}[h]
    \centering
        \begin{tabular}{l l l}
            \hline
            \textbf{Method} & \textbf{Backbone} & \textbf{Angular Error ($^\circ$)$\downarrow$} \\
            \hline
            ETH-XGaze baseline & CNN & 4.5 \\
            FullFace & CNN & 6.53 \\
            GazeCapsNet & Capsule-based & 5.75 \\
            Gaze360 & CNN & 4.46 \\
            SwAT & Transformer & 4.4 \\
            PoolFormer & Transformer & 3.64 \\
            \textbf{CapStARE (Ours)} & \textbf{Capsule-Sequential} & \textbf{3.36} \\
            \hline
        \end{tabular}
    \caption{Within-dataset evaluation on ETH-XGaze under subject-independent protocol.}
    \label{tab:within_ethxgaze}
\end{table}

Table \ref{tab:within_ethxgaze} reports within-dataset performance on ETH-XGaze. CapStARE achieves the lowest angular error ($3.36^\circ$), outperforming all compared CNN-, capsule-, and transformer-based approaches.

\begin{table}[h]
    \centering
        \begin{tabular}{l l l l}
            \hline
            \textbf{Method} &
            \textbf{$D_E \rightarrow D_M$ ($^\circ$)$\downarrow$} &
            \textbf{$D_E \rightarrow D_G$ ($^\circ$)$\downarrow$} &
            \textbf{Sequential Modeling} \\
            \hline
            GazeCapsNet & 12.4 & 15.2 & x \\
            SwAT & 12.1 & 22.9 & Attention-based \\
            Gaze360 & 6.24 & N/A & BiLSTM \\
            \textbf{CapStARE (Ours)} & \textbf{2.65} & \textbf{9.06} & \textbf{GRU} \\
            \hline
        \end{tabular}
    \caption{Cross-dataset zero-shot evaluation using explicit source-to-target notation. Models are trained on ETH-XGaze ($D_E$) and evaluated on MPIIFaceGaze ($D_M$) and Gaze360 ($D_G$) without target-domain fine-tuning.}
    \label{tab:cross_dataset_eth}
\end{table}

Table \ref{tab:cross_dataset_eth} presents cross-dataset transfer results from ETH-XGaze to MPIIFaceGaze and Gaze360. As expected, all methods experience performance degradation under domain shift. Nevertheless, CapStARE achieves the lowest angular error on both MPIIFaceGaze ($2.65^\circ$) and Gaze360 ($9.06^\circ$), indicating strong transfer performance across heterogeneous acquisition conditions. Interestingly, lower absolute angular errors can be observed on MPIIFaceGaze despite the presence of domain shift. This behavior may be partially explained by differences in dataset difficulty, as MPIIFaceGaze contains a narrower gaze-angle range and reduced head-pose variability compared with ETH-XGaze. Therefore, cross-dataset performance should be interpreted not only in terms of absolute error values but also in relation to the intrinsic complexity of the target dataset. Since all competing results are reported from their original publications, minor differences in preprocessing pipelines, backbone capacity, image normalization, and evaluation protocols may affect direct numerical comparability. Therefore, results should be interpreted primarily as an indication of competitive cross-dataset generalization rather than as a strictly controlled head-to-head comparison.

\begin{figure}[t]
\centering
\begin{subfigure}{0.24\textwidth}
    \includegraphics[width=\textwidth]{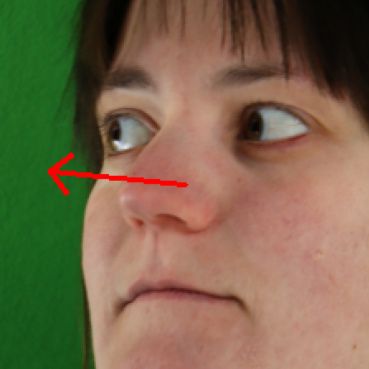}
    \caption{}
\end{subfigure}
\begin{subfigure}{0.24\textwidth}
    \includegraphics[width=\textwidth]{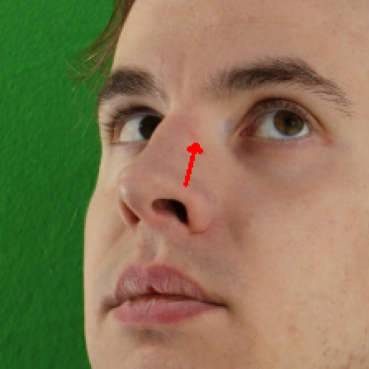}
    \caption{}
\end{subfigure}
\begin{subfigure}{0.24\textwidth}
    \includegraphics[width=\textwidth]{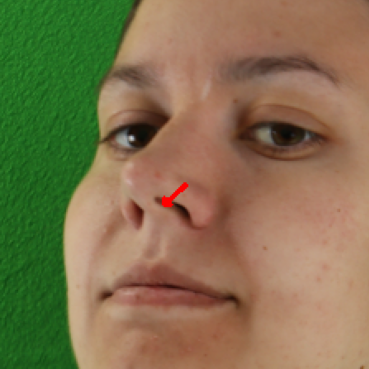}
    \caption{}
\end{subfigure}
\begin{subfigure}{0.24\textwidth}
    \includegraphics[width=\textwidth]{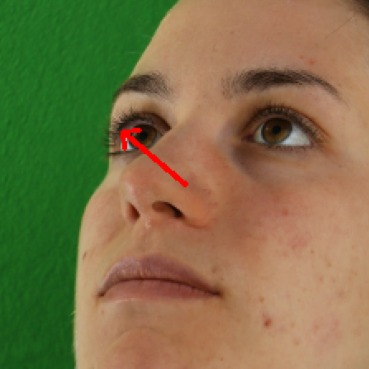}
    \caption{}
\end{subfigure}

\begin{subfigure}{0.24\textwidth}
    \includegraphics[width=\textwidth]{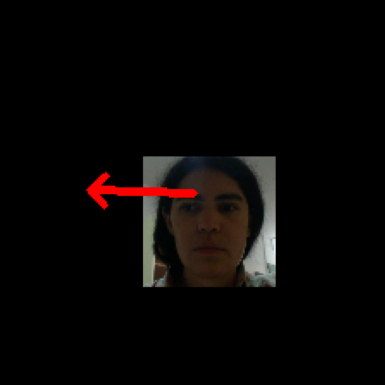}
    \caption{}
\end{subfigure}
\begin{subfigure}{0.24\textwidth}
    \includegraphics[width=\textwidth]{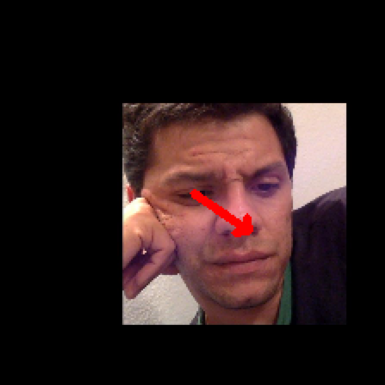}
    \caption{}
\end{subfigure}
\begin{subfigure}{0.24\textwidth}
    \includegraphics[width=\textwidth]{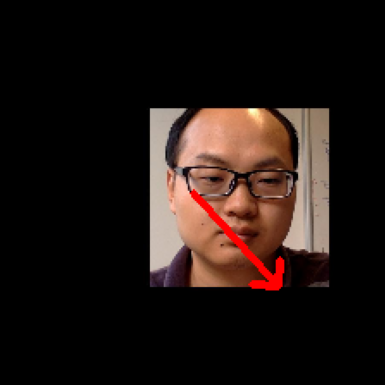}
    \caption{}
\end{subfigure}
\begin{subfigure}{0.24\textwidth}
    \includegraphics[width=\textwidth]{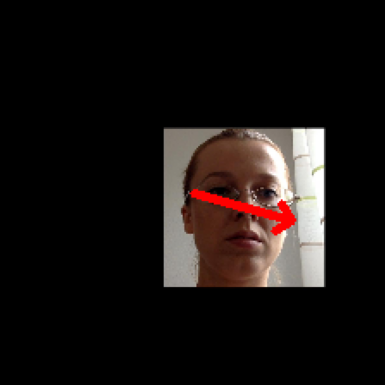}
    \caption{}
\end{subfigure}

\begin{subfigure}{0.24\textwidth}
    \includegraphics[width=\textwidth]{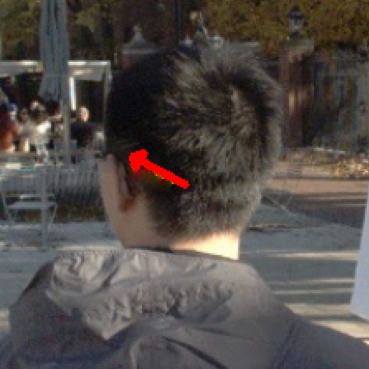}
    \caption{}
\end{subfigure}
\begin{subfigure}{0.24\textwidth}
    \includegraphics[width=\textwidth]{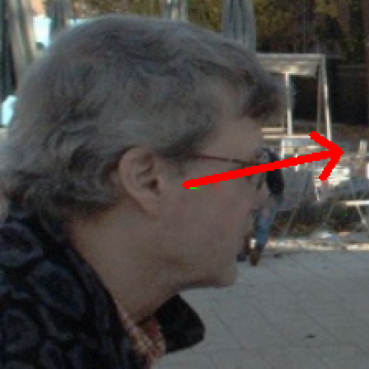}
    \caption{}
\end{subfigure}
\begin{subfigure}{0.24\textwidth}
    \includegraphics[width=\textwidth]{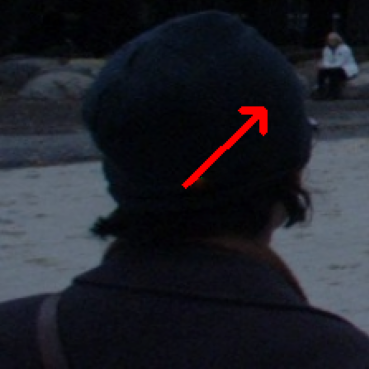}
    \caption{}
\end{subfigure}
\begin{subfigure}{0.24\textwidth}
    \includegraphics[width=\textwidth]{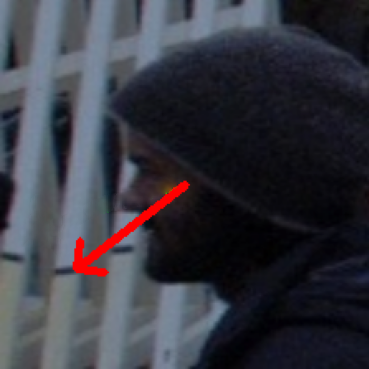}
    \caption{}
\end{subfigure}

\caption{Qualitative results across datasets. (a-d) ETH-XGaze: robustness to extreme head poses. (e-h) MPIIFaceGaze: adaptability to inter-subject variability. (i-l) Gaze360: resilience in unconstrained conditions (lighting, occlusion).}
\label{fig:qualitative_all}
\end{figure}

Figure \ref{fig:qualitative_all} presents representative qualitative examples across ETH-XGaze, MPIIFaceGaze, and Gaze360. The examples illustrate the behavior of CapStARE under challenging conditions, including extreme head poses, illumination changes, partial occlusions, and unconstrained interaction scenarios. Consistent with the quantitative results reported in Tables \ref{tab:within_ethxgaze} and \ref{tab:cross_dataset_eth}, CapStARE appears to maintain relatively stable gaze predictions across substantial appearance and environmental variability. In particular, proposed capsule-sequential formulation appears to preserve structurally relevant facial relationships under partial visibility and pose variation conditions while maintaining robust performance across diverse acquisition settings. Overall, the qualitative observations complement the quantitative benchmarking by illustrating how CapStARE behaves under complex real-world scenarios.

\section{Discussion}

The experimental results consistently suggest that robust appearance-based gaze estimation benefits more from structured representation learning and lightweight contextual reasoning than from increasing architectural scale alone. Across configuration studies, controlled ablations, and benchmarking experiments, performance improvements repeatedly emerge from architectural organization rather than from parameter expansion.

A first observation concerns the importance of structured facial decomposition. Multiple experimental findings support this conclusion. Increasing the number of capsules from four to eight did not improve performance despite doubling the parameter count, while capsule knockout experiments showed that removing any capsule degrades accuracy and that progressively removing multiple capsules causes severe deterioration. Furthermore, disabling capsule formation produced a substantially larger performance drop than removing attention routing alone. Collectively, these results suggest that capsule representations provide an effective structural prior for gaze estimation by preserving localized facial organization throughout the inference process. The qualitative visualizations further support this interpretation, showing consistent specialization toward semantically meaningful facial regions without explicit region-level supervision.

Interestingly, the internal configuration studies also reveal that larger architectures do not necessarily translate into better performance. Similar trends were observed across capsule count, attention head count, and representational dimensionality analyzes. Although larger configurations occasionally achieved marginal improvements, these gains were generally negligible relative to the associated increase in model size and computational cost. These findings indicate that gaze estimation performance saturates once sufficient representational capacity is achieved and that architectural structure becomes more important than further parameter expansion.

The ablation studies additionally highlight the complementary roles of capsules, relational routing, and sequential decoding. While capsule formation provides the strongest individual contribution, attention-based routing further improves performance by enabling interactions between structured facial representations across neighboring observations. The strongest results are consistently obtained when both mechanisms operate jointly, indicating that effective gaze estimation benefits from preserving localized structure while simultaneously allowing contextual interactions between semantically related facial regions.

A second major finding concerns the importance of contextual reasoning beyond isolated appearance cues. Sequence-length experiments demonstrate a consistent reduction in angular error as additional neighboring observations are incorporated. This monotonic improvement indicates that neighboring observations contain complementary information that cannot be fully recovered from static appearance alone. Notably, performance gains begin to saturate as sequence length increases suggesting that relatively short contextual windows already capture most of the useful local continuity available under the evaluated acquisition protocol.

The decoder ablations further reinforce the importance of contextual modeling. Dual-path decoding substantially outperformed a single recurrent pathway despite nearly identical computational complexity, while enforcing weight sharing between decoder branches also produced significant degradation. Together, these findings suggest that contextual information benefits from multiple complementary recurrent representations rather than being compressed into a single latent state. Independent optimization appears to encourage specialization across different contextual dependencies, improving the model's ability to integrate heterogeneous gaze-related cues.

The backbone comparison provides additional insight into the characteristics of gaze estimation as a visual recognition problem. Under identical training conditions, ConvNeXt consistently outperformed EfficientNet, ViT, and Swin Transformer. While these results should not be interpreted as universal backbone rankings, they suggest that preserving local geometric facial structure may be particularly important for appearance-based gaze estimation. Fine-grained periocular cues, subtle facial asymmetries, and head-eye relationships often provide critical information for directional disambiguation. Consequently, locality-preserving inductive priors may remain advantageous under limited gaze-specific adaptation data, even when transformer architectures demonstrate strong performance in broader vision tasks. Similar observations have been reported by ADGaze \citep{LI2025111536}, where convolutional backbones consistently outperformed transformer alternatives under identical gaze estimation protocols, suggesting that effective gaze prediction may depend more strongly on preserving localized geometric structure than on maximizing global token interactions.

Beyond predictive accuracy, the results also highlight the practical efficiency of the proposed framework. Across multiple configuration studies, the most accurate models were often among the most compact alternatives. CapStARE achieves real-time inference with latency below 10 ms while maintaining competitive accuracy across all evaluated datasets. This behavior suggests that robust gaze estimation does not necessarily require large architectures, but rather effective utilization of structured spatial and contextual information. Such characteristics are particularly relevant for deployment oriented applications, including human-computer interaction, assistive technologies, and social robotics, where low latency and computational efficiency remain essential requirements.

Finally, the comparative benchmarking results indicate that the benefits of the proposed formulation extend beyond the source dataset. CapStARE achieved the lowest angular error under both within-dataset and cross-dataset evaluation protocols, including substantial improvements on MPIIFaceGaze and Gaze360 without target-domain fine-tuning. Although all methods experience some degree of performance degradation under domain shift, recent studies have identified appearance bias as a major factor limiting cross-dataset gaze estimation performance, as models often learn subject- and environment-specific visual characteristics that do not generalize across acquisition settings \citep{KIM2024110441}. In this context, the comparatively stable transfer behavior observed in CapStARE suggests that preserving structured facial organization throughout contextual aggregation may partially mitigate such appearance entanglement by encouraging localized specialization rather than globally entangled representations. While additional studies would be required to isolate the precise mechanisms underlying this robustness, the results indicate that structured capsule representations combined with lightweight contextual reasoning provide a promising direction for improving generalization in appearance-based gaze estimation.

Overall, the presented evidence suggests that robust gaze estimation can benefit from combining localized facial decomposition, relational aggregation, and short-horizon contextual modeling within a computationally efficient framework. Rather than relying on increased architectural scale, the proposed approach demonstrates that structured organization of gaze-relevant information can simultaneously improve accuracy, interpretability, efficiency, and cross-domain robustness.

\section{Limitations and Future Work}

Despite the encouraging results obtained across both within-dataset and cross-dataset evaluations, several limitations remain that motivate future research.

First, CapStARE is intentionally designed around short-horizon contextual aggregation rather than unrestricted temporal modeling. Although the sequence-length experiments demonstrate that neighboring observations provide substantial benefits for gaze estimation, the proposed architecture cannot explicitly capture long-range behavioral dependencies or extended interaction dynamics. Consequently, its advantages may be reduced in scenarios involving highly discontinuous observations, abrupt viewpoint transitions, or long-term gaze patterns. Future work could explore hierarchical recurrent architectures, memory-augmented mechanisms, or lightweight transformer modules capable of incorporating longer contextual histories while preserving real-time inference.

Second, the proposed sequence construction protocol relies on locally ordered observations generated from dataset-specific acquisition procedures rather than fully continuous real-world interaction streams. While this design allows controlled evaluation of contextual reasoning on standard appearance-based gaze datasets, additional validation under continuous human-computer and human-robot interaction scenarios would provide a more comprehensive assessment of practical deployment behavior.

Third, although capsule representations improve interpretability through structured facial decomposition, capsule specialization emerges entirely through task optimization and is not explicitly supervised. As a result, the learned regions do not necessarily correspond to anatomically meaningful facial components. Recent studies have shown that incorporating eye landmarks as auxiliary supervision can improve gaze estimation by providing stronger geometric priors and structural guidance during representation learning \citep{SUN2024109980, GOU2024110760}. Similarly, explicit modeling of facial landmarks and head pose has been shown to provide complementary geometric information that can improve facial representation learning and pose awareness \citep{ZOU2025111393}. Future work could therefore investigate the integration of facial landmarks, eye-state information, weak region-level supervision, or anatomical priors into the capsule learning process. Such extensions may further improve robustness, consistency, and interpretability, particularly under severe occlusion, appearance variation, or extreme head-pose conditions.

Another limitation concerns the visual nature of the current framework. CapStARE relies exclusively on RGB appearance information and does not explicitly incorporate geometric, depth, or 3D structural cues. While the cross-dataset experiments indicate strong transfer capability, performance still degrades under challenging conditions involving large pose variations, strong motion blur, or extreme gaze angles. Integrating lightweight geometric reasoning or depth-aware representations may therefore provide additional robustness in highly unconstrained environments.

The cross-dataset experiments further highlight an important open challenge in appearance-based gaze estimation. Although CapStARE achieved the strongest transfer performance among the evaluated approaches, a noticeable performance gap remains between within-dataset and cross-dataset evaluation. This observation suggests that gaze estimation continues to be sensitive to differences in camera geometry, subject appearance, environmental conditions, and acquisition protocols. Recent domain generalization approaches have addressed this problem through contrastive learning and feature disentanglement strategies designed to encourage domain-invariant gaze representations \citep{XIA2025111244}. Incorporating similar mechanisms into the proposed capsule-sequential framework may further improve robustness under heterogeneous acquisition conditions. Future work should additionally investigate self-supervised representation learning and personalization strategies capable of reducing dataset-specific bias while preserving deployment simplicity.

Beyond purely visual modeling, recent multimodal approaches suggest promising directions for future development. Methods such as GazeCLIP \citep{wang2026gazeclip} and gaze-language alignment frameworks \citep{mondal2025gaze} indicate that language-guided supervision may provide additional contextual information beyond visual appearance alone. Similarly, recent architectures have explored semantic conditioning through CLIP-based representations and context-aware domain adaptation mechanisms to improve robustness under heterogeneous acquisition conditions \citep{ZHAO2026116120}. Extending CapStARE toward multimodal or vision-language formulations could improve semantic contextual reasoning, zero-shot transfer capability, and human-centered interaction performance.

Finally, future work should evaluate the proposed framework in real-world deployment scenarios involving unrestricted motion, dynamic environments, and continuous interaction. Such studies would provide valuable insight into the practical robustness, adaptability, and long-term reliability of structured contextual gaze estimation systems operating outside controlled benchmark conditions.

\bibliographystyle{unsrt}
\bibliography{cas-refs}

@INPROCEEDINGS{abdelrahman2022l2csnetfinegrainedgazeestimation,
  author={Abdelrahman, Ahmed A. and Hempel, Thorsten and Khalifa, Aly and Al-Hamadi, Ayoub and Dinges, Laslo},
  booktitle={2023 8th International Conference on Frontiers of Signal Processing (ICFSP)}, 
  title={{L2CS-N}et : Fine-Grained Gaze Estimation in Unconstrained Environments}, 
  year={2023},
  volume={},
  number={},
  pages={98-102},
 note = {\url{https://doi.org/10.1109/ICFSP59764.2023.10372944}}
}

@inproceedings{Zhang_2017,
   title={It’s Written All Over Your Face: Full-Face Appearance-Based Gaze Estimation},
   url={http://dx.doi.org/10.1109/CVPRW.2017.284},
   DOI={10.1109/cvprw.2017.284},
   booktitle={2017 IEEE Conference on Computer Vision and Pattern Recognition Workshops (CVPRW)},
   publisher={IEEE},
   author={Zhang, Xucong and Sugano, Yusuke and Fritz, Mario and Bulling, Andreas},
   year={2017},
   month=jul }

@inproceedings{kellnhofer2019gaze360physicallyunconstrainedgaze,
    author = {Petr Kellnhofer and Adria Recasens and Simon Stent and Wojciech Matusik and Antonio Torralba},
    title = {Gaze360: Physically Unconstrained Gaze Estimation in the Wild},
    booktitle = {IEEE International Conference on Computer Vision (ICCV)},
    month = {October},
    year = {2019}
}

@INPROCEEDINGS{9956687,

  author={Cheng, Yihua and Lu, Feng},

  booktitle={2022 26th International Conference on Pattern Recognition (ICPR)}, 

  title={Gaze Estimation using Transformer}, 

  year={2022},

  volume={},

  number={},

  pages={3341-3347},

  doi={10.1109/ICPR56361.2022.9956687}}

@Article{app13105901,
AUTHOR = {Yan, Chao and Pan, Weiguo and Xu, Cheng and Dai, Songyin and Li, Xuewei},
TITLE = {Gaze Estimation via Strip Pooling and Multi-Criss-Cross Attention Networks},
JOURNAL = {Applied Sciences},
VOLUME = {13},
YEAR = {2023},
NUMBER = {10},
ARTICLE-NUMBER = {5901},
URL = {https://www.mdpi.com/2076-3417/13/10/5901},
ISSN = {2076-3417},
DOI = {10.3390/app13105901}
}

@misc{liu2022convnet2020s,
      title={A {ConvNet} for the 2020s}, 
      author={Zhuang Liu and Hanzi Mao and Chao-Yuan Wu and Christoph Feichtenhofer and Trevor Darrell and Saining Xie},
      year={2022},
      eprint={2201.03545},
      archivePrefix={arXiv},
      primaryClass={cs.CV},
      url={https://arxiv.org/abs/2201.03545}, 
}

@article{article,
author = {Brooks, Rechele and Meltzoff, Andrew},
year = {2005},
month = {12},
pages = {535-43},
title = {The development of gaze following and its relation to language},
volume = {8},
journal = {Developmental science},
doi = {10.1111/j.1467-7687.2005.00445.x}
}

@article{hriAdmoniHenny,
author = {Admoni, Henny and Scassellati, Brian},
year = {2017},
month = {03},
pages = {25},
title = {Social Eye Gaze in Human-Robot Interaction: A Review},
volume = {6},
journal = {Journal of Human-Robot Interaction},
doi = {10.5898/JHRI.6.1.Admoni}
}

@inproceedings{Zhang_2015,
   title={Appearance-based gaze estimation in the wild},
   url={http://dx.doi.org/10.1109/CVPR.2015.7299081},
   DOI={10.1109/cvpr.2015.7299081},
   booktitle={2015 IEEE Conference on Computer Vision and Pattern Recognition (CVPR)},
   publisher={IEEE},
   author={Zhang, Xucong and Sugano, Yusuke and Fritz, Mario and Bulling, Andreas},
   year={2015},
   month=jun, pages={4511–4520} }

@inproceedings{zhang_2020,
author = {Zhang, Xucong and Park, Seonwook and Beeler, Thabo and Bradley, Derek and Tang, Siyu and Hilliges, Otmar},
title = {{ETH-XGaze}: A Large Scale Dataset for Gaze Estimation Under Extreme Head Pose and Gaze Variation},
year = {2020},
isbn = {978-3-030-58557-0},
publisher = {Springer-Verlag},
address = {Berlin, Heidelberg},
booktitle = {Computer Vision – ECCV 2020: 16th European Conference, Glasgow, UK, August 23–28, 2020, Proceedings, Part V},
pages = {365–381},
numpages = {17},
location = {Glasgow, United Kingdom}
}

@book{duchowski2017eye,
  author    = {Duchowski, Andrew T.},
  title     = {Eye Tracking Methodology: Theory and Practice},
  edition   = {3rd},
  publisher = {Springer},
  year      = {2017},
  address   = {Cham},
  isbn      = {978-3-319-57883-5},
  doi       = {10.1007/978-3-319-57883-5}
}

@INPROCEEDINGS {krafka2016eyetracking,
author = { Krafka, Kyle and Khosla, Aditya and Kellnhofer, Petr and Kannan, Harini and Bhandarkar, Suchendra and Matusik, Wojciech and Torralba, Antonio },
booktitle = { 2016 IEEE Conference on Computer Vision and Pattern Recognition (CVPR) },
title = {{ Eye Tracking for Everyone }},
year = {2016},
volume = {},
ISSN = {1063-6919},
pages = {2176-2184},
note = {\url{https://doi.ieeecomputersociety.org/10.1109/CVPR.2016.239}},
publisher = {IEEE Computer Society},
address = {Los Alamitos, CA, USA},
month =Jun}

@INPROCEEDINGS {jindal2024spatiotemporalattentiongaussianprocesses,
author = { Jindal, Swati and Yadav, Mohit and Manduchi, Roberto },
booktitle = { 2024 IEEE/CVF Conference on Computer Vision and Pattern Recognition Workshops (CVPRW) },
title = {{ Spatio-Temporal Attention and Gaussian Processes for Personalized Video Gaze Estimation }},
year = {2024},
volume = {},
ISSN = {},
pages = {604-614},
note = {\url{https://doi.ieeecomputersociety.org/10.1109/CVPRW63382.2024.00065}},
publisher = {IEEE Computer Society},
address = {Los Alamitos, CA, USA},
month =Jun}

@article{byersheinlein2021gaze,
  title = {The development of gaze following in monolingual and bilingual infants: A multi‑laboratory study},
  author = {Byers‑Heinlein, Krista and Tsui, Rachel K.Y. and van Renswoude, Daan and Black, Alexis K.},
  journal = {Infancy},
  volume = {26},
  number = {1},
  pages = {4--38},
  year = {2021},
  doi = {10.1111/infa.12360},
  url = {https://doi.org/10.1111/infa.12360}
}

@article{franchak2022beyond,
  title = {Beyond screen time: Using head-mounted eye tracking to study natural behavior},
  author = {Franchak, John M. and Yu, Chen},
  journal = {Advances in Child Development and Behavior},
  volume = {62},
  pages = {61--91},
  year = {2022},
  doi = {10.1016/bs.acdb.2021.11.001},
  url = {https://doi.org/10.1016/bs.acdb.2021.11.001},
  publisher = {Elsevier}
}

@article{stower2021gaze,
  title={A meta-analysis of children’s trust in social robots},
  author={Stower, Rebecca and Kappas, Arvid and Hoffman, Guy and Leite, Iolanda},
  journal={International Journal of Social Robotics},
  volume={13},
  pages={1979–2001},
  year={2021},
  doi={10.1007/s12369-021-00789-3}
}

@INPROCEEDINGS{10208362,
  author={Wang, Hengfei and Oh, Jun O and Jin Chang, Hyung and Na, Jin Hee and Tae, Minwoo and Zhang, Zhongqun and Choi, Sang-Il},
  booktitle={2023 IEEE/CVF Conference on Computer Vision and Pattern Recognition Workshops (CVPRW)}, 
  title={Gaze{C}aps: Gaze Estimation with Self-Attention-Routed Capsules}, 
  year={2023},
  volume={},
  number={},
  pages={2669-2677},
  doi={10.1109/CVPRW59228.2023.00267}}

@Article{s25041224,
AUTHOR = {Muksimova, Shakhnoza and Valikhujaev, Yakhyokhuja and Umirzakova, Sabina and Baltayev, Jushkin and Cho, Young Im},
TITLE = {Gaze{C}apsNet: A Lightweight Gaze Estimation Framework},
JOURNAL = {Sensors},
VOLUME = {25},
YEAR = {2025},
NUMBER = {4},
ARTICLE-NUMBER = {1224},
URL = {https://www.mdpi.com/1424-8220/25/4/1224},
PubMedID = {40006453},
ISSN = {1424-8220},
DOI = {10.3390/s25041224}
}

@INPROCEEDINGS{8784770,
  author={Zhou, Xiaolong and Lin, Jianing and Jiang, Jiaqi and Chen, Shengyong},
  booktitle={2019 IEEE International Conference on Multimedia and Expo (ICME)}, 
  title={Learning A 3D Gaze Estimator with Improved Itracker Combined with Bidirectional LSTM}, 
  year={2019},
  volume={},
  number={},
  pages={850-855},
  doi={10.1109/ICME.2019.00151}}

@inproceedings{ryan2025gaze,
  title={Gaze-{LLE}: Gaze target estimation via large-scale learned encoders},
  author={Ryan, Fiona and Bati, Ajay and Lee, Sangmin and Bolya, Daniel and Hoffman, Judy and Rehg, James M},
  booktitle={Proceedings of the Computer Vision and Pattern Recognition Conference},
  pages={28874--28884},
  year={2025}
}

@article{karmi2025appearance,
  title={An Appearance-based VisionTransformer Network for Enhanced Gaze Estimation},
  author={Karmi, Rawdha and Mastouri, Rekka and Rahmany, Ines and Khlifa, Nawres},
  journal={Signal, Image and Video Processing},
  volume={19},
  number={9},
  pages={742},
  year={2025},
  publisher={Springer}
}

@article{wang2026eye,
  title={Eye-to-action: Real-time robotic grasping via lightweight gaze estimation},
  author={Wang, Baiyang and Wang, Xujian and Fang, Ming and Li, Hua and Wang, Hongjun},
  journal={Expert Systems with Applications},
  pages={131744},
  year={2026},
  publisher={Elsevier}
}

@article{wang2026gazeclip,
  title={Gaze{CLIP}: Enhancing gaze estimation through text-guided multimodal learning},
  author={Wang, Jun and Ruan, Hao and Wen, Liangjian and Dai, Yong and Wang, Mingjie},
  journal={Neurocomputing},
  pages={133584},
  year={2026},
  publisher={Elsevier}
}

@inproceedings{mondal2025gaze,
  title={Gaze-Language Alignment for Zero-Shot Prediction of Visual Search Targets from Human Gaze Scanpaths},
  author={Mondal, Sounak and Sendhilnathan, Naveen and Zhang, Ting and Liu, Yue and Proulx, Michael and Iuzzolino, Michael Louis and Qin, Chuan and Jonker, Tanya R},
  booktitle={Proceedings of the IEEE/CVF International Conference on Computer Vision},
  pages={2738--2749},
  year={2025}
}

@inproceedings{personnic2026learning,
  title={Learning spatio-temporal feature representations for video-based gaze estimation},
  author={Personnic, Alexandre and B{\^a}ce, Mihai},
  booktitle={Proceedings of the IEEE/CVF Winter Conference on Applications of Computer Vision},
  pages={5121--5130},
  year={2026}
}

@article{xia2025timegazer,
  title={Time{G}azer: Temporal Modeling of Predictive Gaze Stabilization for AR Interaction},
  author={Xia, Yaozheng and Zhu, Zaiping and Pang, Bo and Wang, Shaorong and Li, Sheng},
  journal={arXiv preprint arXiv:2510.01561},
  year={2025}
}

@article{SUN2024109980,
title = {Gaze estimation with semi-supervised eye landmark detection as an auxiliary task},
journal = {Pattern Recognition},
volume = {146},
pages = {109980},
year = {2024},
issn = {0031-3203},
doi = {https://doi.org/10.1016/j.patcog.2023.109980},
url = {https://www.sciencedirect.com/science/article/pii/S0031320323006787},
author = {Yunjia Sun and Jiabei Zeng and Shiguang Shan},
}

@article{GOU2024110760,
title = {Cascaded learning with transformer for simultaneous eye landmark, eye state and gaze estimation},
journal = {Pattern Recognition},
volume = {156},
pages = {110760},
year = {2024},
issn = {0031-3203},
doi = {https://doi.org/10.1016/j.patcog.2024.110760},
url = {https://www.sciencedirect.com/science/article/pii/S0031320324005119},
author = {Chao Gou and Yuezhao Yu and Zipeng Guo and Chen Xiong and Ming Cai},
}

@article{KIM2024110441,
title = {Appearance debiased gaze estimation via stochastic subject-wise adversarial learning},
journal = {Pattern Recognition},
volume = {152},
pages = {110441},
year = {2024},
issn = {0031-3203},
doi = {https://doi.org/10.1016/j.patcog.2024.110441},
url = {https://www.sciencedirect.com/science/article/pii/S0031320324001924},
author = {Suneung Kim and Woo-Jeoung Nam and Seong-Whan Lee},
}

@article{XIA2025111244,
title = {Collaborative contrastive learning for cross-domain gaze estimation},
journal = {Pattern Recognition},
volume = {161},
pages = {111244},
year = {2025},
issn = {0031-3203},
doi = {https://doi.org/10.1016/j.patcog.2024.111244},
url = {https://www.sciencedirect.com/science/article/pii/S0031320324009956},
author = {Lifan Xia and Yong Li and Xin Cai and Zhen Cui and Chunyan Xu and Antoni B. Chan},
}

@article{ZOU2025111393,
title = {Towards unsupervised learning of joint facial landmark detection and head pose estimation},
journal = {Pattern Recognition},
volume = {162},
pages = {111393},
year = {2025},
issn = {0031-3203},
doi = {https://doi.org/10.1016/j.patcog.2025.111393},
url = {https://www.sciencedirect.com/science/article/pii/S0031320325000536},
author = {Zhiming Zou and Dian Jia and Wei Tang},
}

@article{ZHAO2026116120,
title = {GMGaze: MoE-based context-aware gaze estimation with CLIP and multiscale transformer},
journal = {Knowledge-Based Systems},
volume = {345},
pages = {116120},
year = {2026},
issn = {0950-7051},
doi = {https://doi.org/10.1016/j.knosys.2026.116120},
url = {https://www.sciencedirect.com/science/article/pii/S0950705126008464},
author = {Xinyuan Zhao and Yihang Wu and Ahmad Chaddad and Sarah A. Alkhodair and Reem Kateb},
}

@article{LI2025111536,
title = {ADGaze: Anisotropic Gaussian Label Distribution Learning for fine-grained gaze estimation},
journal = {Pattern Recognition},
volume = {164},
pages = {111536},
year = {2025},
issn = {0031-3203},
doi = {https://doi.org/10.1016/j.patcog.2025.111536},
url = {https://www.sciencedirect.com/science/article/pii/S0031320325001967},
author = {Duantengchuan Li and Shutong Wang and Wanli Zhao and Lingyun Kang and Liangshan Dong and Jiazhang Wang and Xiaoguang Wang},
}

@Article{info17030224,
AUTHOR = {Xu, Guanghui and Zhang, Xiaoyang and Zhao, Wanli and Mao, Zhongjie and Li, Yue and Li, Duantengchuan and Dong, Liangshan},
TITLE = {Landmark-Guided Gaze Estimation via Conditional Keypoint Generation and Cross-Attention Fusion},
JOURNAL = {Information},
VOLUME = {17},
YEAR = {2026},
NUMBER = {3},
ARTICLE-NUMBER = {224},
URL = {https://www.mdpi.com/2078-2489/17/3/224},
ISSN = {2078-2489},
DOI = {10.3390/info17030224}
}

@article{ZHAO2026113452,
title = {GMMGaze: Dynamic coarse-to-fine gaze estimation based on gaussian mixture modeling},
journal = {Pattern Recognition},
volume = {178},
pages = {113452},
year = {2026},
issn = {0031-3203},
doi = {https://doi.org/10.1016/j.patcog.2026.113452},
url = {https://www.sciencedirect.com/science/article/pii/S0031320326004188},
author = {Wanli Zhao and Duantengchuan Li and Lingyun Kang and Shutong Wang and Xiaoguang Wang},
}

\end{document}